# Combining Qualitative and Computational Approaches for Literary Analysis of Finnish Novels

Emily Öhman[1]  Riikka Rossi[1]
Waseda University  University of Helsinki
ohman@waseda.jp  riikka.rossi@helsinki.fi

## Abstract

What can we learn from the classics of Finnish literature by using computational emotion analysis? This article tries to answer this question by examining how computational methods of sentiment analysis can be used in the study of literary works in conjunction with a qualitative or more 'traditional' approach to literature and affect. We present and develop a simple but robust computational approach of affect analysis that uses a carefully curated emotion lexicon adapted to Finnish turn-of-the-century literary texts combined with word embeddings to map out the semantic emotional spaces of seminal works of Finnish literature. We focus our qualitative analysis on selected case studies: four works by Juhani Aho, Minna Canth, Maria Jotuni and F. E. Sillanpää, but provide emotion arcs for a total of 975 Finnish novels.

We argue that a computational analysis of a text's lexicon can be valuable in evaluating the large distribution of the emotional valence in a text and provide guidelines to help other researchers replicate our findings. We show that computational approaches have a place in traditional studies on affect in literature as a support tool for close-reading-based analyses, but also allowing for large-scale comparison between for example, genres or national canons.

## Introduction

The study of literature provides interesting insights for an interdisciplinary study of emotion since literature can be considered a genre in which the affective functions of language are of principal importance (Hogan 2011, 1). The research on literature and emotions is a flourishing and rapidly developing subfield within literary studies, which has in recent years gained new impetus through the development of cognitive science, philosophy, and history of emotions as well as the theory of affect[2]. Recent affect-related research topics range from the study of literature and empathy (Keen 2007) to the study of literature and cognition (Hogan 2011), negative affect and tone in texts (Ngai 2005) to empirical perspectives (Sklar 2013; Van Lissa et al. 2018).

There is a need to develop new methods which would combine the often-separated empirical study and qualitative text analysis. Although the idea that domain-experts should be fully utilized in interdisciplinary research is by no means novel (see e.g. Tangherlinin & Leonard, 2013) in the context of Finnish language and literature especially, but in literary analysis in general too, the setting of this paper is unique. We hope that our method of combining expertise from several disciplines and working together to analyze the results will lead the way for true interdisciplinary collaboration where both fields are advanced by utilizing computational methods to extract either previously undiscovered knowledge or knowledge at a previously impossible scale. Previous studies have shown that in interdisciplinary research, it is typically the computationally trained participant who both creates the model and performs the analysis of the results (Bartlett 2018, 3). In our approach, we want to emphasize the importance of the domain expert and the computational expert working together at all steps of the process, especially in the model design or approach and the results analysis.

---

[1] Both authors contributed equally.
[2] We may speak of an "affective turn" in literary studies of the 21th century. The affective power of literary texts is known since Aristotle's *Poetics*, but in the 20th century, many schools such as formalism, new criticism, structuralism and post-structuralism orientated the attention to formal and structural aspects of texts, whereas the study of emotions was excluded and considered as susceptible to researchers' subjective emotions.

Our study explores the use of quantitative tools for literary studies. Using such quantitative or computational tools for literary analysis is not novel in itself and indeed, a whole subfield of digital humanities and natural language processing often referred to as *computational literary studies* (CLS) exists. We note that CLS as a field has been accused of either producing results that show the obvious or alternatively producing results that are not robust enough to withstand scrutiny (Da 2019, 601). The truth is probably not quite as harsh, but there does seem to be a trend in CLS where the studies are interesting from a computational perspective but offer little or no new knowledge or tools to traditional literary scholars. Our aim is to showcase simple methods that can be used with a qualitative analysis to gather supporting evidence for traditional theoretical frameworks. In our case, we focus on the study of literature and emotions, specifically mood through affective expressions. This study is the first of several where we explore these concepts by combining qualitative and quantitative approaches.

We focus on using variations of text analytics in general and emotion detection, a more fine-grained type of sentiment analysis[3], specifically in combination with affective qualitative analyses of Finnish literature conducted by domain experts. Sentiment analysis has been used in its current form for decades and has become a particularly active field within language technology in the past decade (for an overview see e.g., Mäntylä et al. 2018). There are many different approaches, however, all current approaches can be traced back to either lexicon-based, data-driven, or hybrid methods in which lexicons can be used to enhance the data for machine learning approaches or, for example, as a part of word embeddings[4].

We hypothesize that (1) computational methods such as emotion detection or emotion recognition can successfully be used for mood detection in Finnish novels to support a qualitative analysis and use a priori qualitative interpretations of select texts to compare the quantitative results against. Furthermore, (2) we posit that traditional descriptions of an author's style are evident in the quantitative output. Additionally, our paper provides emotion arcs for 975 classic Finnish novels as open data. Our experimental setup includes qualitative analyses of key novels that are used as a baseline to which we compare the results from our computational approach to prove hypothesis (1) and show the usefulness of the 975-book corpus with emotion arc annotations. In our analysis and validation process we also show that our second hypothesis (2) holds true.

---

[3] As it is understood in the field of language technology and natural language processing.
[4] Word embeddings are vector space representations of the semantic mappings of language. Words close in distance cover similar semantic spaces and words that carry similar meanings in context are grouped near each other in vector space.

**Background and Previous Work**
The study of affect and emotions in Finnish literature has flourished during the past decade. Recent studies have mapped the historical poetics of emotions in the work of Finnish authors (Maijala 2008; Kurikka 2013; Hollsten and Helle 2016) and in the context of specific literary genres and period-styles (Isomaa 2012, Rossi 2020). They have also contributed to the general theory of literature and emotions especially regarding the notion of tone and emotional effects (Lyytikäinen 2017) and charted culture-specific emotions and affects and their representation in Finnish literature (Nykänen and Rossi 2020). Recent studies also include and discuss empirical approaches to the study of literature and emotion (e.g. Rossi 2021, 2023; Lahtinen & Löytty 2022).

While the study of emotions has become a rapidly developing approach to Finnish literature, methods in computational *sentiment analysis* are almost nonexistent in the field[5] apart from our recent work (Rossi & Öhman 2021 & 2022; Rossi 2022) and some meta discussions (Capková, 2021). In its simplest form sentiment analysis means automatically assigning a text (or other modality) either valency (positive/negative/neutral) or an emotion category (sadness/anger/…) or both. In affective sciences, the term *emotional valence* is used as a dimensional variable to describe the positive or negative character of emotions, of components of the emotional response such as subjective feelings or behavioral responses, and of emotion eliciting stimuli (Brosch and Moors 2009, 401-402). Accordingly, the study of valence in texts means measuring the scale of positive and negative affect in language[6].

By the term ***affect***, we refer to an overarching sense of a mental state that is characterized by a feeling in comparison with rational thinking (e.g. Clore 2009). There is much debate about what emotions are and how they relate to affect both between different fields and within fields as well and much confusion around the terminology too (Munezero et al., 2014). In this work we consider the relationship between affect and emotion to be such that affect is subjectively experienced emotions, which in turn are intentional and often targeted projections of feelings. We acknowledge that the term "emotion" is one of the fuzziest in all of the sciences.

Computational literary studies focusing on affect often concentrate on data-mining emotion words in text sources (see e.g. Kim & Klinger 2018 or Yadahalli et al. 2017). For example, Ashok et al. (2013) showed that the fewer sentiment-laden words a novel contained, the more successful it was likely to be in terms of sales, and Alm and Sproat (2005) showed that fairy tales begin neutral and end happy, and Mohammad (2012) showed how books portray different entities through co-occurring emotion words. More recent work by the Aarhus University DIGHUMLAB has focused on the correlation between affective content measured by fractal sentiment arcs and the success of novels (Hu et al., 2021, Bizzoni et al., 2022). Additionally, Baunvig et al., (2020) used exclamation marks as markers of emotion and intent in the writings by Grundtvig. Although the exact methodology varies, these studies focus largely on the emotional lexicon in the studied works of literature. However, it is important to note that although emotion-associated words are an important component in constructing affective narratives, they are only a part of how emotions and affect are expressed in text (Kövecses 2000, 2). This is especially true in genres that use specific emotive text tools, including word choice, to evoke specific affective states in the reader. For instance, in our earlier study on affect in Finnish authors we observed that one author, Maria Jotuni, who uses many positive emotion words, is heavily ironic in her description of emotions (Öhman & Rossi 2021). The ironic tone in her works actually converts the positive affects into negative: "love" usually signals the opposite: lack of love. A list of emotion words in her works would thus not be informative alone but needs to be complemented by the study of affective valence in general and an understanding of the genre, and authors' style. For this reason, our methods include measuring *the emotional intensity* and *valence* of each emotion-associated word in a text, that is, we not only measure whether a word is associated with an emotion but also how intense that emotion association is.

---

[5] Computational literary studies focusing on affect or emotion is a thriving field, here we are talking specifically about using computational methods in traditional literary analysis.
[6] Many commonly used datasets are based on review data where stars or similar metrics are used to determine whether a review is positive, neutral, or negative. Modern machine learning algorithms often take advantage of large language models such as BERT (Devlin et al. 2018) to understand language in context and therefore also classify language data better in downstream tasks such as emotion detection and classification in specific domains such as legal texts or medical texts. Although such models work well on texts from some domains, others require more training of the model to make it work well on that specific domain (certain words have completely different connotations in different fields: "jargon"). Such training requires computational skills but also access to GPU power or similar powerful computer equipment, the use and acquisition of which can be expensive.

The lexicon-based analysis can be criticized for not taking context into account to the same extent as machine learning models supported by large language models do, sometimes excluding even valence shifters. However, the semantic level is only one aspect of language. From a phenomenological perspective, a semantically negative sentence can appear positive to the reader. Take, for instance the phrase: "*alavilla mailla hallan vaara*", which has been referred to as the most beautiful phrase in the Finnish language; the repetition of the so-called pleasant phonemes (l) creates an almost poetic image in the reader's mind.[7] However, semantically, the meaning of the phrase is much more negative, as it indicates a danger of frost in low fields, and thus potential crop failures (i.e. connotations of hunger and death in the history Finland).

Although several studies that compared lexicon-based sentiment analysis to data-driven approaches have concluded that the data-driven ones are more accurate (see e.g. van Attenveldt 2020), there are concerns about how such comparisons are conducted (Öhman 2021, Laaksonen et al. 2023) and the comparison data has exclusively consisted of shorter texts such as tweets or news headlines. Such texts are short and therefore even when normalizing a text (calculating an emotion intensity value for the whole text per word count) if just one word is wrong, the entire result is worthless. Similarly, if just one word is missing from the lexicon, it has a big impact on the overall results. When dealing with larger texts, such problems are greatly diminished and only categorical misassociation of words (e.g. an erroneous emotion association in the lexicon for the particular domain) will impact the overall score whereas a single misassociation will have very little impact on the overall data (see e.g. Teodorescu & Mohammad 2022 for a thorough discussion on the accuracy of lexicon-based sentiment analysis). We can also use statistical significance testing to ensure that comparative results are truly meaningful (see e.g. Koljonen et al. 2022).

**Data and Methods**

In this section, we present an overview of the data in quantitative terms and discuss the methods we used in detail. Text mining and lexical analysis for affect

---

[7] On phonosymbolism and pleasantness of l, see e.g. Whissel 1999.

45Some of the simplest and most robust methods to investigate the content of books computationally involve basic corpus linguistic inquiries. We can determine basic facts such as type-to-token ratios between different books to see if one author uses more complex/varied language than another and examine the use of specific phrases with the help of part-of-speech tagging to investigate the use of specific syntactic structures. This approach also allows for the extraction of, for example, adjectives to investigate the direct ways the author describes various entities in their prose.

These approaches are perhaps not always useful to literary scholars, but can serve as a base when starting a qualitative study and as a simple way of getting examples of specific expressions one is interested in. However, once we have these basics, we can start exploring how they relate to emotions by utilizing emotion lexicons, word embeddings, and language models with oor without machine learning, among other quantitative tools.

**Dataset creation and preprocessing**

The data (the literary works and their metadata) were collected from Project Gutenberg and preprocessed with a Finnish version of *chapterize* both to remove the legal texts in the beginning and end of the text files and later to examine chapter specific affective language use (see Öhman and Rossi 2022 for more details). The size of our total corpus is 975 comprising the first 1000 Finnish language books available on Project Gutenberg that are actually encoded in utf-8 (25 were discarded due to encoding issues). We present emotion arcs for all of the works, but in this paper focus our attention on four specific literary texts. The works are mainly from the late 1800s and early 1900s with only a handful of exceptions and cover various genres but can for the most part be considered as having been written in Modern Finnish which is said to have been well-established by 1870 if not earlier (Ikola 1965, 39). We used regular expressions to extract the title, author, publication year, as well as whether the book had been originally written in Finnish or not. After experimenting with different lemmatizers[8] and text normalization ("translating" non-standard words to standard language) we chose to lemmatize the texts using the Turku Neural Parser (Kanerva et al. 2021). The final data consists of 2,938,032 sentences and 41,417,116 tokens[9].

**Editing the Lexicon**

We use the Finnish Emotion Intensity Lexicon (FEIL) as a starting point in our work (Öhman 2022). FEIL is based on the NRC Emotion Intensity Lexicon (Mohammad & Kiritchenko 2018). The lexicon was originally created by annotating English data using crowdsourcing and best-worst scaling, and then translated to 106 new languages using Google Translate with few or no manual corrections. In FEIL these translations have been checked and revised, both in terms of duplicates, as well as connotations and other semantic inconsistencies. The further revisions we did for FEIL are discussed in the Data & Method section of this paper as well as in Öhman & Rossi (2022,2023). The lexicon uses Plutchik's wheel of emotions (Plutchik 1980) as a basis with the emotions *anger, anticipation, disgust, fear, joy, sadness,* and *trust* with the words linked to one or more of these emotions with a value between 0 and 1 representing the intensity of the emotion for that specific word.

---

[8] A tokenizer splits a text into word-like units. A lemmatizer yields the base, or dictionary, form of a token.
[9] Tokens are often equivalent with words, but typically also contain punctuation and other similar characters which are counted as their own tokens.

*Accepted in Scandinavian Studies Journal, issue 97.3 (2025). DOI and link will be added once available.*

FEIL is a lexicon created based on the contemporary meaning of the words as well as their contemporary emotion associations. The original NRC lexicon was furthermore annotated by mostly North Americans in the late noughties. The North American cultural associations of some words, religious words in particular, are naturally different from the Finnish connotations even with perfect translations. As the annotations were collected recently, there are also global cultural shifts that have taken place between the time of publication of the books in our collection and the annotations. This meant that we needed to make sure that the semantic shift of words had not drifted too far from the time period of the selected works and that the emotion associations made sense from a cultural perspective as well.

In the process of revising FEIL many instances of cultural connotative differences were addressed, and the lexicon revised for turn-of-the-century Finnish literature. Öhman (2022, 426-427) uses the original English words *pious*, *devout*, *saintly*, and *godly* as examples of over-generalization as these four words had all been automatically translated as *hurskas* by Google Translate. In this case, it was possible to find better translations for all terms without adjusting the emotion intensity.

**Table 1. Examples of over generalization in machine translation**

| Original English | Google Translate | Edited Translation |
| --- | --- | --- |
| pious | hurskas | hurskas |
| devout | hurskas | harras |
| saintly | hurskas | pyhimysmäinen |
| godly | hurskas | jumalinen |

To ensure domain- and period-suitability we extracted two lists from the original results. The first one contained the most common words in the texts and the second one contained the most common emotion associated words from the texts. For the first list we examined the 2000 most common words for words that we felt should be in the lexicon as they had a clear emotion-association but were not. To help add these words to the lexicon with the correct and objective intensity values and emotion associations we used a word vector approach (word2vec, Mikolov et al., 2017) where we created a vector space model based on the words in context in all 975 books, and then with the help of cosine similarity look up the words with the closest semantic space match and assigned that word's emotion association and intensity values to the word that was missing from the lexicon. In this way we could make sure our biases would not affect the lexicon, and that the emotion associations would be culturally relevant and as objective as possible (for related work see e.g., Maas et al. 2011; Yu et al. 2017; Ye et al. 2018; Diegoli & Öhman 2024; Öhman et al. 2024). The number of words that were added to the lexicon was relatively small, so we were able to manually check that all associations made sense particularly for the specific domain in question. This led to the association for, e.g., the words *kirkas (clear)*, *valkoinen (white)*, and *valkea (white, bright)* to be identical. We also had to add verb forms of some very common emotion associated words such as "rakastaa" (to love). In English the dictionary entry for the verb and noun form are typically identical (to/a love), but in Finnish these have different word forms (rakastaa/rakkaus) and so for the most part only the noun form had been present in the lexicon before we added the missing verb forms.

**Table 2. Examples of deleted, added, and common emotion-associated words in the data.**

| Deleted words | Added words | Common emotion-associated words |
|---|---|---|
| saada (receive, auxilliary) | kylmä (cold) | kiinni (closed, shut) |
| käydä (go, be in operation) | pyhä (holy, sacred) | rikkominen (the act of breaking sth.) |
| puhua (to talk) | pimeä (dark) | eksyä (to get lost) |
| itse (self, reflexive) | suloinen (sweet, cute) | romahtaa (to collapse) |
| aika (time, quite) | heikko (weak) | likainen (dirty) |

As for the second list containing the most commonly emotion-associated words, we ended up removing some of the most common words as we felt they had no emotion association at all. Among these words was "saada" (to receive) as it is used in many different grammatical constructs, similar to auxiliary verbs, where it does not carry the meaning it does as a standalone word. We also removed a few words due to ambiguity, e.g. "aika" can mean both "time" and "quite" and is therefore a potential valence shifter which we wanted to avoid having in the lexicon as an independent lemma. In future work, such words might be reintroduced to the lexicon by utilizing trigrams and part-of-speech tags.

The final lexicon had a fairly balanced distribution of emotion words with *fear*, *trust*, *joy*, and *anger* between 1176-1560 each, and *anticipation* and *disgust* at 832 and 951 respectively. *Sadness* is the only outlier at only 207 instances. The correlation (or co-annotation) between *fear* and *sadness* is remarkably high; the negative emotions tend to co-occur, but for *sadness* 71% of *sadness*-associated words are also associated with *fear*. The same is not as true for the positive emotions, which are less likely to co-occur with other positive emotions than negative emotions are with the exception of *anticipation* which co-occurs with *joy* and *trust*, but also *fear*.

**Results**

In this section we present an overview of the quantitative results.

**Emotion Intensity in Selected Works**

There is a sweet spot in the amount of data lexicon-based methods work best for. Too short, and a single word might really affect the distribution beyond its real-world impact. Too long, and eventually the text will be so varied in terms of emotional expressions that the distribution will converge on the distribution in the lexicon. Our texts are long enough for individual words not having much of an impact, but short enough that the distribution of words remains relevant and distinct between authors and texts.

Using the intensity lexicon FEIL yielded the results in table 3. These are the emotion-word distributions for association and intensity. The number is the sum of all word intensity scores in a book normalized as a score per 10,000 words to enable comparison of different length texts. *Hurskas Kurjuus* clearly contains the most and the most intense emotion words particularly negative emotion words (although *anticipation* is usually considered a positive emotion, words that relate to *fear*, tend to also be associated with *anticipation* so *anticipation* on its own can be either positive or negative). It is also the only novel that clearly expresses *sadness* lexically to any real extent.

Jotuni's novel is known for its positive outlook on life and has been referred to as peaceful and optimistic. This is also reflected in the emotion word distribution with the highest proportional intensities of *joy* and *trust* associated words. It should be noted that Jotuni is heavily ironic in her description of emotions. Despite this, contemporary critics agree with the overall results shown in table 3 as discussed in previous sections.

*Sadness* is the least common emotion in the lexicon even after adjusting the lexicon to reflect the emotions associated with all the most common words in our texts. Alm & Sproat (2005, 673) found that *surprise* was expressed mostly by surrounding sentences, i.e. that *surprise* was not simply expressed in one sequence, but with contrasting sequences and was therefore difficult to detect computationally. We suggest something similar is going on with our texts in terms of *sadness*. However, they found that *anger* and *sadness* were both more likely to be preceded and followed by the same emotion, something that we have not seen to the same extent in our data.

**Table 3. Emotion intensity sums by word count in each novel.**

| Title | Word count | Anger | Anticip. | Disgust | Fear | Joy | Sadn. | Trust |
|---|---|---|---|---|---|---|---|---|
| Kauppa-Lopo | 12068 | 65.96 | 142.60 | 47.76 | 80.13 | 132.20 | 26.94 | 158.82 |
| Rautatie | 28097 | 40.81 | 135.38 | 26.68 | 54.73 | 91.72 | 8.60 | 133.82 |
| Hurskas Kurjuus | 52382 | **88.64** | **206.80** | **71.10** | **131.83** | 164.37 | **53.04** | 189.12 |
| Arkielämää | 27292 | 66.64 | 156.76 | 48.75 | 93.35 | **193.47** | 30.86 | **221.41** |

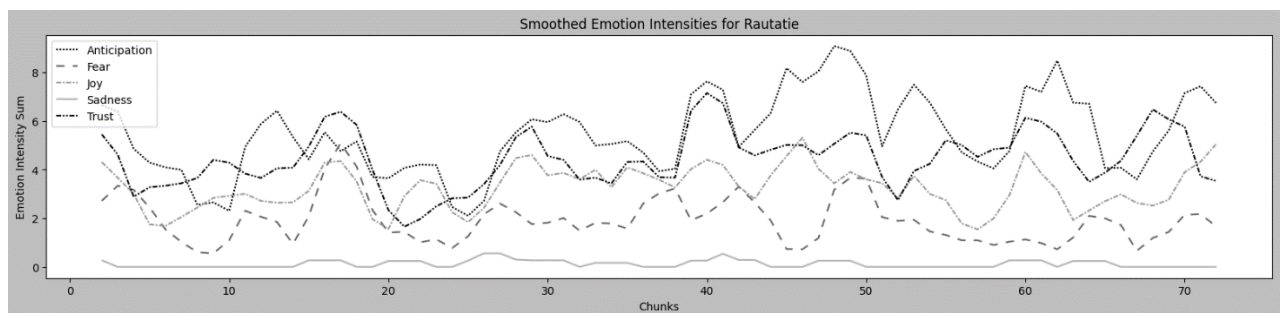

**Figure 1. Smoothed (select) emotion intensities for Aho's *Rautatie* (The Railroad).**

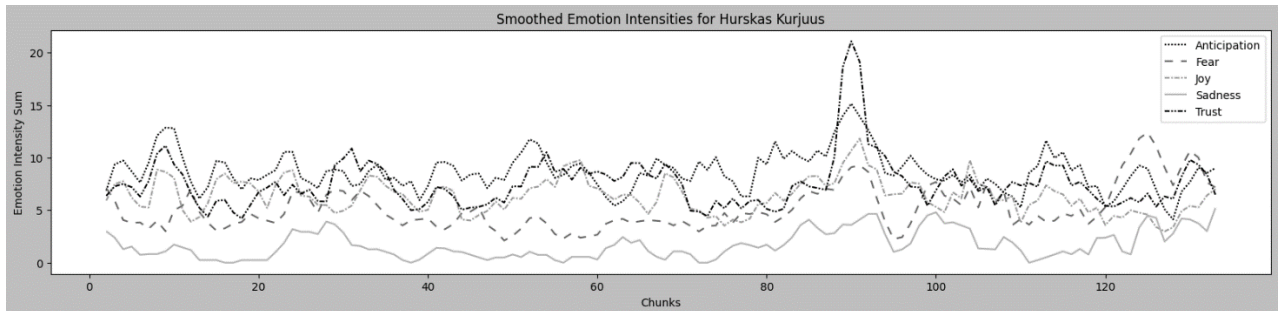

**Figure 2. Smoothed (select) emotion intensities for Sillanpää's *Hurkas Kurjuus* (Sacred Misery).**

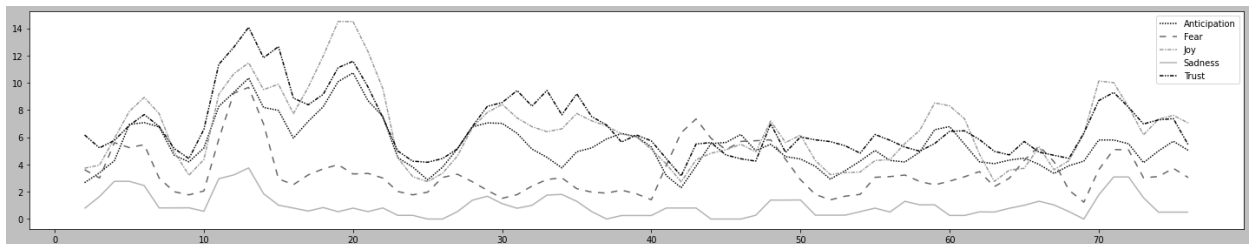

**Figure 3. Smoothed (select) emotion intensities for Jotuni's *Arkielämää* (Everyday life).**

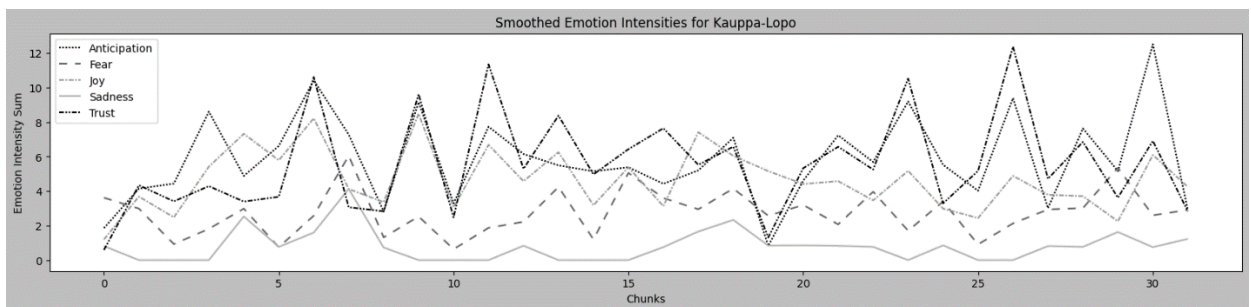

**Figure 4. Smoothed (select) emotion intensities for Canth's *Kauppa-Lopo*.**

**Evaluation of Results**

In this section we present the results of the computational analysis and compare it with the qualitative analysis.

**Juhani Aho:** ***The Railroad***

Juhani Aho's *Rautatie, eli, Kertomus ukosta ja akasta, jotka eivät olleet sitä ennen nähneet* (1884, tr. in English as *The Railroad or, A Tale of an Old Man and an Old Woman Who Had Never Seen It Before*) is considered as one of Aho's most important works. It describes the effects of modernity on ordinary people's lives. The novel tells the story of an older couple, Matti and Liisa, living in a remote forest village, and how they embark on their first railway journey. While new technology arouses curiosity, the encountering of modernity turns into deception. During the journey Matti ends up intoxicated, they miss their destination, and finally the couple must walk back home. However, the novel concludes with a fairy-tale-like happy ending and suggests a continuation of life in the forest idyll.

Previous research on *Rautatie* has paid attention to melancholia and nostalgia evoked by the encountering of modernity and the separation from the premodern idyllic lifestyle. For instance, in his study on Aho's humor Unto Kupiainen (1937, 27) considers the basic mood of *The Railroad* as "lightly melancholic" despite the comic elements and humorous effects of the story. Nostalgia and melancholia are characteristic of Aho's works in general (e.g. Rossi 2020, 141).

*Sadness*, however, is not highlighted in our quantitative results of emotion intensities in Aho's *Rautatie*. As previously discussed, lexicon-based measuring of sadness is challenging, since *sadness* typically co-occurs with other emotions and contouring words. This might reflect the nature of sadness itself, which can be considered as an emotion of low intensity, sometimes as a blue mood of long duration, caused by loss or felt when things are permanently unattainable (Ekman & Davidson 1994, 24). The disperse, all-encompassing nature is salient Aho's work, in which the indirect evocation of *sadness* plays an important role. Milieu descriptions reflect the melancholic mood. For instance, the very beginning of *Rautatie* describes a winter landscape, frigid cold and then focuses on a lonely, hungry magpie in the yard tree. The feeling of sorrow is then explicitly mentioned in the text, yet it is projected to the bird: "The magpie's mood turns sad. It is hungry and cold." (Aho 2012, 20).[10]

The emotion of *fear* is more explicit in the quantitative analysis of Aho's novel. The results reflect the depiction of the character's emotional experience and especially the *fear* triggered by encroaching modernity. For instance, the computational analysis indicates an increase of emotional intensity in chapter two (chunks 16-18), in a scene in which the male protagonist, Matti, imagines the railroad and anticipates its potential threat. The breakthrough of modernity evokes *fear*, even a sense of the end of the world. Especially the Biblical vocabulary occurring in the narrated monologue on Matti's experience raises the intensity of the lexicon tracked computationally: *maailmanloppu* (apocalypse) and *piru* (the Devil), verbs such as *huutaa* (yell), *manata* (curse) and *leimahtaa* (sparkle) occurring in these descriptions evoke urgency, force and abstract concepts and raise the energy level of the text. The scene happens at night, and during the heavy frost, Matti travels with his horse and sleigh. The darkness during the sleighride, the coldness, and the movement in the semantic level of the narrative match the computational reading of the text.

While it is charged with great emotional intensity, it is important to note that this scene does not represent a turning point in the narrative of the story. The difference between important and less notable events cannot be judged based on the lexicon alone, but it requires an evaluation of the significance of the events and persons in the narrative continuum. Sometimes meaningful events can even hide themselves in narrative gaps (ellipsis), and in these cases a lexical-based analysis would not be sufficient alone.

The computational analysis, however, matches the increased intensity of a scene considered as a peak event in the narrative: Chapter VII, which depicts the first railway experience of Matti and Liisa. At the moment of the train's departure the protagonists grip their seats; the sudden bellringing of the departure frightens Liisa. From the protagonists' perspective, the movement feels like a horse would be running wildly; a sense of falling and disintegration are mentioned in 11the text. The computational analysis matches this increase of intensity by connecting higher intensity words such as *hajota* (disintegrate), *hätä* (emergency), *uskaltaa* (to have courage).

---

[10] The original in Finnish: "Harakan mieli käy surulliseksi, sillä on nälkä ja vilu …" Aho 1884, 3.

Despite the protagonists' disappointing encounter with modernity, Aho's *Rautatie* ends with positive feelings. The return to a nature paradise is presented as possible. The final scene of the novel is reminiscent of the happy ending of fairy-tales. Also, the turn towards the positive is in some respects reflected in the computational reading which shows a slight rise in *joy* in the end: words such as *aurinko* (the sun), *koti* (home), and *taivas* (heaven) evoke *joy*. The sky is clear, the sun is shining again, and the happiness of nature is paired with the protagonists' cheerful feelings. The idyllic description of nature and the narrator's compassionate view of animals and human beings soften the critical view to modernization.

**Maria Jotuni:** ***Everyday life***

Maria Jotuni's *Arkielämää* (Everyday life 1909) is a short novel, or novella, which describes the life of ordinary rural people on a single summer day in a remote countryside village.[11] The episodic text affords a view of the life of a collective through a male protagonist, a solitary vagabond called Nyman, referred to in the novel as "priest", who frequently visits the village. He meets the villagers and listens to their joys and sorrows, in particular those of ordinary peasant women. They all stand at the threshold of life-long choices in love and marriage. The one-day timespan covers pivotal events of human life from death to birth, and through the flashbacks to the villagers' past, painful events are evoked: Nyman meets, for instance, an infanticidal mother; and he is also told about an incest case recently revealed in the village.

While the narrative events include descriptions of *unhappy love*, *anxiety*, *sadness*, the novel's mood has been interpreted as positive. "Her approach is dispassionately peaceful and reconciliatory like life itself" (Kaila 1910, 122). "The view to life is positive and optimistic", (Koskenniemi 1909, cit. Niemi 2001, 102). "With respect to unity of tone and style, there are not many works that equate to Jotuni's novel" (Koskelainen 1909, 127). A sense of sacredness is prominent in the novel (Rossi 2010; 2020, 153-155). Through the male protagonist, Jotuni's work presents a sense of connection and even ecstatic experiences of nature.

The computational analysis provides further evidence on the positive mood. The intensity of Jotuni's text is relatively smooth and the emotion of *joy* is highlighted (see table 3). In general, the measured intensity of Jotuni's novel is lower than the emotional intensities of *Rautatie*, *Hurskas kurjuus* and *Kauppa-Lopo*, (see table 3).

From the perspective of narrative events, the computational analysis indicates an increase of emotional intensity in the first part of the novel (1-25). The peak of emotions is salient in the third chapter (12-15), where Nyman meets the infanticidal mother who thinks back the traumatic events of the past. She is particularly worried if her unbaptized baby has ascended into Heaven or got to hell. Explicit emotion words might heighten the intensity: *rakkaus* (love), *epätoivo* (despair) and *tuska* (pain) are included in the lexicon of this scene. Religious vocabulary charged with emotions is in ample use in this scene too: *taivas* (Heaven), *synti,* (sin), *kadotus,* (doom), *armo* (mercy).

---

[11] On Jotuni's work as a representative of the Nordic modern breakthrough, see Rossi 2010.

However, like in Aho's work, emotional intensity does not necessarily equate to narrative intensity. The infanticidal mother is a side character whose experience is presented as an example of the villagers as a whole. The evaluation of larger semantic units and their role in the narrative continuum still requires human analysis.

**Sillanpää:** *Sacred Misery*

Frans Emil Sillanpää's *Hurskas kurjuus* (tr. as Meek Heritage, literally "Sacred Misery" 1919) analyzes Finland's path to the Civil War (1918) by depicting an individual life-story, a poor tenant farmer, Juha, who ends up a Red Guard soldier. By looking at the war from the perspective of the losing Reds it challenges the victory of the Whites. Negative emotions are salient emotional effects of the novel, which describes its protagonist as not only as an object of *disgust* and hatred of the Whites but also rejected by his own class and his Red comrades-in-arms (see Rajala 1983, Rossi 2020). A kind of collective guilt is evoked in the novel, which evolves into a narrative of scapegoating and crucifixion: Sillanpää's anti-hero is an innocent victim, who is executed for a crime that he has not committed.

The strong negative emotions are reflected in the computational analysis, which suggests a high proportion of *anger* and *disgust* in the opening scene. The novel begins with a shocking prologue of the White terror: the protagonist is executed at a mass grave. The computational analysis of the opening scene tracks several words which relate to these shocking events. *Anger*, *disgust* and *fear* are highlighted in the machine-based analysis which associates these emotions to words such as *kurjuus* (misery), *kuolema* (death), *vastenmielinen*, (disgusting) *uhkaava*, (threatening), *kapinallinen* (rebellious).

Sillanpää's novel is rich in accurate depictions of poverty and misery focus on filth and contamination in primitive living conditions: they move from spoiled food to putrefaction, from the sufferings of a crippled child to lethal diseases spreading in Juha's family. The emotion of *disgust* is then used to illustrate the protagonist's existential disgust with life: Juha experiences life as distasteful, sickening, a "sour and dull substance" (Sillanpää 1919, 140). The narrative betrays his feelings of aversion toward life and the world in general.

The description of negative emotions culminates in a chapter entitled "Death does its best" which depicts a series of Death in Juha's family. The computational analysis also suggests a peak of emotions in this chapter (the chunks 88-94). The description of pain and death of Juha's child, who is crippled in an accident is charged with words related to *anger*, *anticipation*, *disgust*, *fear*; in the same way the negatively associated words are salient in the scene narrating Juha's wife cancer and death: *anger*, *disgust*, *fear*.

In contrast, the high emotional intensity in these scenes includes a high proportion of *joy*-related words in these scenes. The *joy*-related words to refer to words such as *kiitollinen* (thankful), *lapsi* (a child), *elämä* (life), *äiti* (mother).

**Canth:** *Kauppa-Lopo*

Minna Canth's *Kauppa-Lopo* (1889) is a naturalistic, tragic story of poverty and illness. The novella presents us a female anti-hero, Lopo, a poor outcast of society, who has drifted into vagabonding, drinking, and petty crimes. Canth's work was inspired by French naturalism and contemporary psychopathological theories (e.g. Rossi 2007, 73; Maijala 2008, 227-230). However, in her work the naturalistic effects of shock and ugliness are used to raise critical discussion on social inequality[12]. The protagonist is described as a victim of poverty and unknown instincts. The narrative underlines her good-hearted and compassionate towards other people, and Canth's narrative invites the reader' to feel sympathy for the poor woman. Instead, the novella unveils the upper classes' contempt and disgust for her, and casts doubt on the morals of the rich and the powerful. Perhaps due to the uncomfortable truths that the novella presents about the upper class, the story was met with anger in the contemporary audience; the critics did not value her social criticism. (Rossi 2007, 52).

Like in the previous examples analyzed above, the quantitative analysis on Canth's work reflects the emotions evoked and thus provides further evidence on the affective aspects of the narrative. For instance, we may look at the intensity of negative emotions, the beginning of Canth's novella, set in prison. The scene underlines the protagonist's ugly appearance, as seen through the prison-guard: "Lopo turned her puffy face to the room and grimaced. Riitta wondered about her ugliness once again. The corners of the mouth and the nasal base were covered with snuff and phlegm, hair tangled, concealing eyes."[13] In the computational analysis, words such as *vankeus* (imprisonment), *varkaus* (theft), *rumuus*; (ugliness), *lima* (phlegm) are connected to negative emotions and thus further demonstrate the negative emotional effects created Canth's text.

At the same time, the overall computational analysis on Canth's narrative is more difficult to interpret than the other case studies. The computational analysis of the opening mood of Canth's text did detect *fear*, *suspense* and *disgust* related to the prison environment and the protagonist's ugly appearance, that is underlined in the description. However, the overall analysis of the word distribution proved not reliable, as the digital reading emphasized positive emotions of *joy* and *trust* although the effects of *fear* and *sadness* are central to Canth's tragic story. *The quantitative* analysis, which indicates positive emotional valence of the lexicon, can also be challenged; yet the passion that is characteristic of Canth's works (see Maijala 2008), can be considered as energizing and in a sense positive.

*The quantified valence measures* signal to a rise of positive emotions towards the ending, although the ending of Canth's story, on the level of narrative, is clearly negative. The protagonist is first described as an object of blame and reproach by other people, then captured for a theft again and the whole narrative ends with an inconsolable cry: "But Lopo felt as if she had been taken to the grave alive. And she cried, —cried more bitter tears than ever before."[14] One explanation lies in the distribution of emotion words analyzed quantitatively: the quantitative analysis suggests that there is a strong emphasis on words that are affiliated to the emotion of *trust*. Moreover, although the ending is clearly negative from the perspective of the protagonist, the shock effect is targeted to positively activate the reader and even add to the reader's sympathy towards the character[15].

---

[12] In these novels, negative emotions are represented with the tragic and serious, naturalist tone. The genre schemata entail affective forecasting and serve to pre-adjust the readers' emotional responses (Menninghaus et al. 2017). For instance, exposure to disgusting matters is likely to be perceived differently in a comedy than a tragedy (Menninghaus. et al 2017).

[13] The original in Finnish: "Lopo käänsi turpeat kasvonsa huoneesen päin ja virnutti. Riitta taaskin ihmetteli hänen rumuuttaan. Suupielet ja nenän-alusta nuuskaisessa limassa, hiukset takussa ja silmillä." (Canth 1889, 50).

[14] The original in Finnish: "Mutta Loposta tuntui kuin olisi häntä elävänä hautaan viety. Ja hän itki, — itki katkerampia kyyneliä, kuin milloinkaan ennen." (Canth 1889, 96).

The challenges in computational analysis of Canth's text perhaps reflect the fact that the lexicon in Canth's work is scarce in comparison with other examples. Description of milieu, which usually requires a broad vocabulary, plays a secondary role in Canth's prose fiction. Instead, her style builds on dialogue narration and the narrated monologue of the protagonists' thoughts and emotions. As a playwright Canth knew how to build dramatic tension and effects through dialogue and inner speech.

**Conclusion**

In conclusion, our study demonstrates that a quantitative study on the lexicon can provide supporting evidence on the evocation on emotions in literary texts. We suggest that especially the study of the general affective atmosphere, or the *mood* of a text, can benefit from a computational study of a text's lexicon. In literary texts, the levels of affective aspects, emotional intensities and various emotional effects contribute to an all-encompassing whole, usually called mood or tone (see. Lyytikäinen 2017; Rossi 2020).[16] Due to its diffuse nature, studying mood is challenging. Namely it is not easy to exactly discern the exact distribution of positive and negative emotions in the text-continuum since the tone of the text is not reducible to a single expression of emotions or to clear-cut units.

However, the lexicon-based analysis can provide a supporting tool for the study of literary moods. The analysis of Aho's *Rautatie* and Jotuni's *Arkielämää* points in this direction. *The quantitative representation of* Aho's work confirms the impression of the melancholic background mood of the novella. The analysis on emotion intensity (table 3) is an interesting result: it suggests a low intensity of emotion-word distribution in Aho's work, thus providing further evidence for previous observations on Aho's style, which scholars have characterized as "smooth" or "even" (Kupiainen 1937, 23).

In the same way, the quantitative analysis provides supporting evidence for the qualitative evaluation of the positive mood in Jotuni's *Everyday life*. Although Jotuni's narrative focuses on unhappiness and loss, events that in real life are likely to arouse unpleasant emotions, the analysis of the valence of the lexicon refers to positive emotions: in the lexicon, words linked to emotions of *joy* and *trust* are foregrounded (table 3). Thus, the quantitative results confirm the critics' and scholars' evaluation of the positive or peaceful tone of the novel. Moreover, the results *implicated by our quantitative approach* (figure 3) support the critic's evaluation on the positive emotional valence of the beginning and the ending of Jotuni's work.

---

[15]This on the contrary, could be viewed as being in line with Plutchik's theory of active emotions such as anger as positive.

[16] Existing studies have already pointed out several aspects which contribute to the creation of a mood, including rhetoric and stylistic choices; narrative perspective; rhythm, tropes and vocabulary, all of which influence on the shaping of the tone of a text and serve as textual cues for the reader in the perception of the mood.

An unexpected finding relates to religious vocabulary. Although emotions themselves and how they are experienced are fairly universal, specific topics tend to have culturally bound valence and even emotion associations. Thus, the more religious nature of the North American annotators of the original lexicon seems to have more accurately predicted the valence of late 1800s Finnish society than what would perhaps been possible had the lexicon originally been annotated by modern Finns. This is evidenced by the positive associations of religious words and the accurate emotion arcs of passages with religious content that does not correspond to a modern anachronistic reading but is more in line with the authors' intended emotion evocation.

The quantitative analysis provides evidence also for the strong negative emotions that have previously been attached to Sillanpää's *Sacred Misery* and point to intensive negative emotions: the lexicon latches on to the highly salient emotions of *disgust*, *anger*, *sadness* and *fear*. The quantitative mood analysis of the opening chapter describing the protagonist's ugliness and death by execution is in accordance with the qualitative evaluation of the emotional effects of *disgust*, *fear,* and *anger*.

At the same time, the comparison of qualitative and quantitative analysis indicates that existing sentiment analysis methods still include many challenges and require further development. Especially tools suitable for a fine-grained analysis of the perspective to subject wait to be developed. While our study demonstrates the computational analysis is largely in line with a qualitative reading of the lexicon intensity, a nuanced view to the interpretation of the narrative requires a human analysis: the most intensive scene is not necessarily the most significant in the narrative. There is thus a need to develop methods which would differentiate lexicon related to, for instance, the main and the side characters, the main and the side episodes in the plot (although these may overlap in different types of narratives). Machine learning methods such as named entity recognition combined with computational lexicon analysis could provide potential solutions.

Besides the abovementioned needs for improvement, other questions need to be solved. One problem with the existing methods in sentiment analysis is that they are frequently based on either Ekman's universal emotions (Ekman, 1971) or Plutchik's (1982) model of basic emotions, both of which were developed to describe human emotional states "in the wild", not for text mining. For example, Dodds et al. (2021) go one step further and aptly point out that valence, arousal, and dominance are not orthogonal categories and that there is a bias towards "safe" words in human language. Simply put, human interpretations of emotional content in texts do not align well with the theoretical models of emotions that are commonly used. As an example, Plutchik's wheel of emotions does not include self-reflective emotions such as shame and guilt, which are considered essential emotions that are limited to the human species and powerfully affected by socio-cultural factors (e.g., Scherer & Wallbot 1994). Historians of emotions have analyzed forms of shame typical of Finnish culture (e.g., Siltala 1994). Indeed, many works in our corpus depict feelings of shame in direct or indirect ways. In future studies, we hope to develop our own emotion classification system better suited for the exploration and analysis of literature, something that has been tangentially explored by e.g., Demszky et al. (2020) and Öhman (2020).

Despite these reservations, we believe that new digital tools developed in conjunction with qualitative analysis and up-to-date theories of literature and emotions, can provide valuable support-tools for studying literary history and permit to compare large groups of texts such as national canons. In particular, the study of the tone and mood of a text can benefit from a quantitative text analysis: as discussed above, the quantitative analysis can provide accurate information on the distribution of positive and negative valence in a text and thus help in evaluating which emotional effects are foregrounded in a text. There are also plenty of opportunities for a big data study in which quantitative analysis can function as a depth finder for a qualitative close-reading; perhaps by utilizing other novel approaches such as fractal sentiment arcs (Hu et al. 2021). For instance, there has been a debate on pessimism and optimism in Nordic and French literature: some have argued that Nordic nineteenth-century literature is more optimistic than French naturalism of the era, while others suggest the contrary (see Rossi 2007). Could a quantitative and contrasting analysis of the emotional valence in a sample of Nordic and French texts shed a new light on this question? We think so and aim to investigate this in future work.

In this paper we have shown that computational approaches to mood detection can be a useful tool in supporting literary analyses of novels. The methods need to be further developed to better handle the quirks and eccentricities of literary writing and better balance the wide range of styles of different authors. However, even in its current stage, the quantitative results support the qualitative analysis and vice versa. The authors' styles are reflected in the quantitative results with emotion arcs generally corresponding to a qualitative reading of the novels in terms of mood, but even more so to the specific affect-evoking vocabulary and style used by each author.

In future work, we hope to further explore the connection between affect and mood and fine-tune our methodology to detect the mood of a literary text even more accurately and thus contribute to the understanding of how mood is constructed in literature.

**Acknowledgements**
This work was supported by JSPS KAKENHI Grant Numbers JP22K18154 and JP24K21058 and the Research Council of Finland grant 354218.